\documentclass[10pt,a4paper]{article}

\usepackage[latin1]{inputenc}
\usepackage{amsmath}
\newcommand{\norm}[1]{\left\lVert#1\right\rVert}

\usepackage{amsfonts}
\usepackage{amssymb}
\usepackage{multicol}
\usepackage{graphicx}
\usepackage{booktabs}
\usepackage{float}
\usepackage{xcolor}
\usepackage{soul}
\usepackage{graphicx}
\usepackage{subcaption}
\usepackage{comment}
\usepackage[symbol]{footmisc}
\usepackage[misc,geometry]{ifsym} 

\usepackage[a4paper, left=2cm, right=2cm, top=2.5cm, bottom=2.5cm]{geometry}
\usepackage{titling}
\usepackage{algorithm}
\usepackage{algpseudocode}
\usepackage[sort,comma,numbers]{natbib}
\usepackage{tikz}
\usepackage{amsmath}
\providecommand{\abs}[1]{\lvert#1\rvert}
\usepackage{authblk}

\renewcommand\Authfont{\fontsize{10}{14.4}\selectfont}
\renewcommand\Affilfont{\fontsize{9}{10.8}\itshape}

\begin{document}
\author[1$\#$]{Cuong Phuc Ngo}
\author[1$\#$]{Amadeus Aristo Winarto}
\author[2,3]{Connie Khor Li Kou}
\author[2]{Sojeong Park}
\author[2]{Farhan~Akram}
\author[2]{Hwee Kuan Lee \thanks{Corresponding author: leehk@bii.a-star.edu.sg}}

\affil[1]{Hwa Chong Institution (College Section)}
\affil[2]{Bioinformatics Institute, Agency for Science, Technology and Research (A*STAR)}
\affil[3]{School of Computing, National University of Singapore}

\affil[$\#$]{Equal contribution}

\renewcommand\Authfont{\fontsize{11}{14.4}\selectfont}
\renewcommand\Affilfont{\fontsize{9}{10.8}\itshape}

\title{Fence GAN: Towards Better Anomaly Detection}
\maketitle

\begin{abstract}
\renewcommand{\thefootnote}{\fnsymbol{footnote}}

\noindent
Anomaly detection is a classical problem where the aim is to detect anomalous data that do not belong to the normal data distribution. Current state-of-the-art methods for anomaly detection on complex high-dimensional data are based on the generative adversarial network (GAN). However, the traditional GAN loss is not directly aligned with the anomaly detection objective: it encourages the distribution of the generated samples to overlap with the real data and so the resulting discriminator has been found to be ineffective as an anomaly detector. In this paper, we propose simple modifications to the GAN loss such that the generated samples lie at the boundary of the real data distribution. With our modified GAN loss, our anomaly detection method, called Fence GAN (FGAN), directly uses the discriminator score as an anomaly threshold. Our experimental results using the MNIST, CIFAR10 and KDD99 datasets show that Fence GAN yields the best anomaly classification accuracy compared to state-of-the-art methods.
%1. anomaly detection is important task
%2. current state-of-the-art methods on anomaly detection for complex high-dim images use GAN
%3. But normal GAN objective encourages distribution of generated samples to overlap with real data, which is not %directly aligned with the anomaly detection objective, and hence people find the resulting discriminator is ineffective for anomaly detector
%4. Our contribution: We propose a simple tweak to the GAN objective such that generated samples lie at the boundary of real data distribution, then use the discriminator score as anomaly score
%5. Experiments show improvement over current best GAN-based methods

\end{abstract}

\section{Introduction}
\noindent
\\Anomaly detection is a well-known problem in artificial intelligence where one aims to identify anomalous instances that do not belong to the normal data distribution \cite{chandola2009anomaly,hawkins1980identification}. It is used in a wide range of applications such as network intrusion \cite{garcia2009anomaly}, credit card fraud \cite{srivastava2008credit}, crowd surveillance \cite{mahadevan2010anomaly, ravanbakhsh2017training}, healthcare \cite{schlegl2017unsupervised} and many more. Traditional classifiers trained in a supervised setting do not work well in anomaly detection since the anomalous data is usually unavailable or very few. Hence, anomaly detectors are usually trained in an unsupervised setting where the distribution of the normal data is learned and instances that are unlikely to be under this distribution are identified as anomalous.

\noindent
\\For complex high-dimensional datasets such as images, traditional methods for anomaly detection are unsuitable. Instead, recent methods based on generative adversarial networks (GANs) have shown state-of-the-art anomaly detection performance by exploiting GANs ability to model high-dimensional data distributions. However, we identify a shortcoming of current GAN-based anomaly detection methods: the usual GAN objective encourages the distribution of generated samples to overlap with the real data, and this is not directly aligned with the anomaly detection objective. The resulting discriminator has been found to be ineffective in detecting anomalous data. Hence, in this paper, we propose a simple modification to the GAN objective such that the generated samples lie at the boundary of real data distribution instead of overlapping it. Our method, which we call Fence GAN (FGAN), trains in the usual adversarial manner with the modified objective and we show that the resulting discriminator can be used as an anomaly detector. We conducted experiments on MNIST, CIFAR10 and KDD99 datasets and show that FGAN outperforms state-of-the-art methods at anomaly detection.

\noindent
\\The main contributions of this paper are as follows:
\begin{itemize}
    \item We propose an anomaly detection method using the basic GAN architecture and framework.
    \item We modify the GAN loss such that the samples are generated only at the boundary of the data distribution unlike the traditional GANs that generate samples over the whole data distribution.
    \item The proposed method is tested on MNIST, CIFAR10 and KDD99 datasets, showing improved accuracies over other state-of-the-art methods.
\end{itemize}

\section{Related work}
\noindent
\\Traditional methods for anomaly detection include one-class SVM \cite{scholkopf2001estimating}, nearest neighbor \cite{eskin2002geometric}, clustering \cite{smith2002clustering}, kernel density estimation \cite{nicolau2016one} and hidden markov models \cite{gornitz2015hidden}. However, such methods are not suitable for high-dimensional image data. Recent developments in deep learning have led to significant progress in supervised learning tasks on complex image datasets \cite{he2016deep, ronneberger2015u}. For anomaly detection, deep learning based methods include deep belief networks \cite{erfani2016high}, variational autoencoders \cite{an2015variational, xu2018unsupervised} and adversarial autoencoders \cite{zhai2016deep,leveau2017adversarial,beggel2019robust, lim2018doping}.

\noindent
\\Among the deep learning methods, generative adversarial networks (GANs) \cite{goodfellow2014generative, radford2015unsupervised} have been the subject of extensive research as they show state-of-the-art performance in modeling complex high-dimensional image distributions. Similarly, GANs have been used for anomaly detection. In AnoGAN \cite{schlegl2017unsupervised}, the authors propose an anomaly detector where the GAN is first trained on the normal images, and for a test image, the latent space is iteratively searched to find the latent vector that best reconstructs the test image. The anomaly score is a combination of the reconstruction loss and the loss between the intermediate discriminator feature of the test image and the reconstructed image. A similar framework is used in ADGAN \cite{deecke2018anomaly}, where the anomaly score is based only on reconstruction loss, the search in latent space is repeated with multiple seeds and both the latent vector and generator are optimized. A more recent method, called Efficient GAN \cite{zenati2018efficient}, makes use of the BiGAN model that is able to map from the image to latent space without iterative search, resulting in superior anomaly detection performance and faster test times. Finally, in the GANomaly framework \cite{akcay2018ganomaly}, the generator consists of encoder-decoder-encoder subnetworks and the anomaly score is based on a combination of encoding, reconstruction and feature matching losses. GANomaly has shown superior performance compared to AnoGAN and Efficient GAN on several image datasets such as MNIST and CIFAR10.

\noindent
\\Except for GANomaly, the GAN-based anomaly detection methods above train GAN with the usual minimax loss function where the generator aims to generate samples that overlap with the data distribution. Under the usual GAN loss function, the discriminator probability score was found to be ineffective \cite{deecke2018anomaly}, and we hypothesize that this is because the discriminator is not explicitly trained to fence the boundary of the data distribution. Contrary to these methods, our proposed Fence GAN aims to learn the boundary of the normal data distribution. We achieve this by modifying the generator's objective to aim to generate data lying on the boundary of the normal data distribution, instead of overlapping with the data distribution. At test time, the anomaly score is simply the discriminator score given to the input data. Our alternative generator objective is similar to the one in \citet{dai2017good}, where they show that for the discriminator to be a good classifier, the generator has to produce complement samples instead of matching with the true data distribution. With the modified GAN loss, Fence GAN does not need to rely on reconstruction loss from the generator and does not require modifications to the basic GAN architecture unlike Efficient GAN and GANomaly. 
\section{Method}
\subsection{Original GAN loss function}
\noindent
\\In the original generative adversarial network by \citet{GAN}, for a set $\mathcal{X}$ of $N$ number of data points $\mathcal{X} = \{\boldsymbol{x}_1, \boldsymbol{x}_2,..., \boldsymbol{x}_N\}$ with $\boldsymbol{x}_i$ in a Euclidean data space $\mathbb{R}^d, \: d \in \mathbb{Z}^+$, $i=1,2,..,N$, which is sampled from a data distribution $p_{data}:\mathbb{R}^d \rightarrow \mathbb{R}^+$, we seek to map points from a prior noise distribution $p_{noise} :\mathbb{R}^k \rightarrow \mathbb{R}^+, \:k \in \mathbb{Z}^+$ to $p_{data}$. For example, if each data point represents an image, then $d$ would be the number of pixels in the image. The dimension $k$ is set arbitrarily.

\noindent
\\The mapping from $p_{noise}$ to $p_{data}$ is done by first using a differentiable function, represented by a ``generator'' multilayer perceptron $G_{\theta}$ with $\theta$ being its weights and biases, to map $p_{noise}$ to the generated distribution $p_g :\mathbb{R}^d \rightarrow \mathbb{R}^+$ from the output $G_{\theta}(\boldsymbol{z}), G_{\theta}(\boldsymbol{z}) \in \mathbb{R}^d$ of $G_{\theta}$ and $\boldsymbol{z}$ is drawn from $p_{noise}$. In addition, we also have a ``discriminator'' multilayer perceptron $D_{\phi} $ with $\phi$ being its weights and biases, which outputs a real value $D_{\phi}(\boldsymbol{x})\in [0, 1]$ that represents the probability of $\boldsymbol{x}$ - a point in $\mathbb{R}^d$ being drawn from $p_{data}$ rather than from $p_g$. $D_{\phi}$ and $G_{\theta}$ engage in a two-player minimax game, with $D_\phi$ and $G_\theta$ being alternatingly trained to minimize their respective loss functions as follows:
\begin{align}
    \mathcal{L}_{G_\theta}^{GAN}(G_\theta,D_\phi,\mathcal{Z}) &= \frac{1}{N}\sum_{i=1}^{N}\Bigg[\log(1- D_{\phi}(G_{\theta}(\boldsymbol{z}_i)))\Bigg]\\
    \mathcal{L}_{D_\phi}^{GAN}(G_\theta,D_\phi,\mathcal{X},\mathcal{Z}) &= \frac{1}{N}\sum_{i=1}^{N}\Bigg[-\log(D_{\phi}(\boldsymbol{x}_i)) -\log(1-D_{\phi}(G_{\theta}(\boldsymbol{z}_i)))\Bigg]
\end{align}

\noindent
\\where $\mathcal{Z} = \{\boldsymbol{z}_1, \boldsymbol{z}_2, ..., \boldsymbol{z}_N\}$ is sampled from $p_{noise}$. $\mathcal{L}_{G_\theta}^{GAN}$ is the loss function of $G_\theta$ and $\mathcal{L}_{D_\phi}^{GAN}$ is the loss function of $D_\phi$.

\noindent
\\In this way, $D_{\phi}$ is trained to differentiate whether $\boldsymbol{x}$ is drawn from $p_{data}$ or from $p_g$. Meanwhile, $G_\theta$ is trained to map $p_{noise}$ to $p_g$ so as to maximise the score of its generated points as given by the discriminator, that is, $D_\phi(G_\theta(\boldsymbol{z}))$.

\noindent
\\The training of GAN is completed if the distribution $p_g$ is indistinguishable from $p_{data}$. When this occurs, $p_{data}$ is estimated by $p_g$. Therefore, the mapping of points from $p_{noise}$ to $p_g$, which is represented by $G_\theta $, is also the mapping of points from $p_{noise}$ to $p_{data}$.

\subsection{Modified loss functions}
\noindent
\\We propose Fence GAN (FGAN) which has a different objective from the original GAN's. Whereas the original GAN aims to generate $p_g = p_{data}$, that is, to generate points at regions of high data density, our objective is to generate points around the boundary of $\mathcal{X}$, which we denote as $\delta\mathcal{X}$. This will enable our discriminator, at the end of training, to draw a boundary ``tightly'' around $\mathcal{X}$. Such a discriminator can then be used as a one-class classifier or an anomaly detector.

\noindent
\\Learning $\delta\mathcal{X}$ directly is known to be an extremely difficult problem in high dimensions \citep{convexhull}. Thus, we use the discriminator score to define the domain of $\delta\mathcal{X}$ and then estimate $\delta\mathcal{X}$ using the generator in FGAN. The generated points $G_{\theta}(\boldsymbol{z})$ then must enclose the real data points tightly as shown in Figure \ref{fig:2Dgauss}(A). In order to achieve our objective, we propose a series of modifications to the loss functions for the generator and discriminator: encirclement and dispersion losses for generator, and weighted discriminator loss for discriminator.

\subsubsection{Generator}
\paragraph{Encirclement Loss}
\noindent
\\In our proposed FGAN, we want the generator to generate points $G_{\theta}(\boldsymbol{z})$ that lie in $\delta\mathcal{X}$. We first define points on $\delta\mathcal{X}$ to be those that yield a discriminator score of $\alpha$. To reflect this, the objective function of the generator in FGAN to be minimized is therefore: 

\begin{align}
    EL\big(G_\theta,D_\phi, \mathcal{Z}\big) &=  \frac{1}{N}\sum_{i=1}^{N}\Bigg[\log(\abs{\alpha-D_{\phi}(G_{\theta}(\boldsymbol{z}_i))})\Bigg]
\end{align}

\noindent
\\where $\alpha\in (0, 1)$ is used for the generator to generate points on $\delta\mathcal{X}$. The rationale for Eq. (3) is that points generated inside of $\mathcal{X}$ will have a discriminator score higher than $\alpha$ and hence the generator will be penalised. On the other hand, points generated far from $\mathcal{X}$ will have a discriminator score less than $\alpha$ and hence the generator is also penalised. Only when points are generated at the $\alpha$-level set of the discriminator score will they yield optimal generator loss. This level set should ideally tightly enclose the real data points. In our experiments, we tune the value for $\alpha$ as a hyperparameter.

\paragraph{Dispersion loss}
\noindent
\\Based on the encirclement loss alone, however, there is no guarantee that the generated points will cover the entirety of $\delta\mathcal{X}$, it may only cover a small part of it, as shown in Figure \ref{fig:2Dgauss}(C). We note that this is similar to the mode collapse problem in GAN. The dispersion loss, which maximizes distance of the generated data points from their centre of mass $\boldsymbol{\mu}$, $\boldsymbol{\mu} \in \mathbb{R}^d$, is thus introduced to encourage the generated points to cover the whole boundary.

\begin{align}
\boldsymbol{\mu} = (\mu_1, \mu_2, ... , \mu_d) \nonumber
\\\boldsymbol{\mu} = \frac{1}{N}\sum_{i=1}^{N} G_{\theta}(\boldsymbol{z}_i)\nonumber
\end{align}

\noindent
\\The dispersion loss is thus:
\begin{align}
\ DL(G_{\theta},\mathcal{Z}) = \frac{1}{\frac{1}{N}{\sum_{i=1}^{N}(\norm{G_{\theta}(\boldsymbol{z}_i) - \boldsymbol{\mu}}_2)}}
\end{align}

\noindent
\\We use $L2$ distance because in our experiments we found that it works better compared to $L1$ and $L\infty$ distances. The loss function of the generator in FGAN to be minimized is defined as the weighted sum of the encirclement loss and the dispersion loss:
\begin{align}
        \mathcal{L}_{generator}^{FGAN}\big(G_\theta,D_\phi,\mathcal{Z}\big) &= EL\:+ \beta \times \:DL \nonumber
        \\&= \frac{1}{N}\sum_{i=1}^{N}\Bigg[\log\Big[\abs{\alpha-D_{\phi}(G_{\theta}(\boldsymbol{z}_i))}\Big]\Bigg] + \beta \times \frac{1}{\frac{1}{N}{\sum_{i=1}^{N}(\norm{G_{\theta}(\boldsymbol{z}_i) - \boldsymbol{\mu}}_2)}}
\end{align}
\noindent
\\where $\beta$ is the dispersion hyperparameter with $\beta \in \mathbb{R}^+$.

\subsubsection{Discriminator}
\paragraph{Weighted Discriminator Loss} 
\noindent
\\As the generator becomes better in approximating $\delta\mathcal{X}$, the discriminator faces a trade-off: to classify real data correctly or classify generated data correctly. If the discriminator focuses more on classifying generated data correctly, then the discriminator will start to classify real data as generated data. Thus, the loss function of the discriminator should be modified to prioritise classifying real data correctly:
\begin{align}
    \mathcal{L}_{discriminator}^{FGAN}\big(G_\theta,D_\phi,\mathcal{X},\mathcal{Z}\big) &= \frac{1}{N}\sum_{i=1}^{N}\Bigg[-\log\big(D_{\phi}(\boldsymbol{x}_i)\big) -\gamma\log\bigg(1-D_{\phi}(G_{\theta}(\boldsymbol{z}_i)\big)\bigg)\Bigg]
\end{align}
\noindent
\\where $\gamma$ is the anomaly hyperparameter with $\gamma \in (0,1]$. When $\gamma$ is less than 1, the discriminator will focus more on classifying the real data points correctly, thus its decision boundary is less likely to bend into the domain of $\mathcal{X}$, allowing the generator to better estimate $\delta\mathcal{X}$. We empirically tune $\gamma$ for each dataset.

\subsection{Fence GAN (FGAN)}
\noindent
\\FGAN is composed of a generator and a discriminator being trained one after another like a typical GAN. The number of steps to train $G_\theta$ and $D_\phi$ for each iteration are hyperparameters to be tuned. However for simplicity, we train both networks once in each iteration.

\begin{algorithm}[H]
  \caption{Stochastic gradient descent training of FGAN.}
  \label{Algorithm}
  \begin{algorithmic}
    \For{\text{number of training iterations}}
        	\State $\bullet$ Sample noise samples $\mathcal{Z} = \{\boldsymbol{z}_1, \boldsymbol{z}_2, ..., \boldsymbol{z}_N\}$ from prior $p_{noise}$
        	\State $\bullet$ Update the generator's parameters:
        	\begin{equation*}
    				\theta \gets \theta - \eta_g \nabla_\theta \mathcal{L}_{generator}^{FGAN}\big(G_\theta,D_\phi, \mathcal{Z}\big)
			\end{equation*}
			
        	\State $\bullet$ Resample noise samples $\mathcal{Z} = \{\boldsymbol{z}_1, \boldsymbol{z}_2, ..., \boldsymbol{z}_N\}$ from  prior $p_{noise}$
        	\State $\bullet$ Sample data samples $\mathcal{X} = \{\boldsymbol{x}_1, \boldsymbol{x}_2, ..., \boldsymbol{x}_N\}$ from real data  distribution $p_{data}$
        	\State $\bullet$ Update the discriminator's parameters:
        	\begin{equation*}
    			\phi \gets \phi - \eta_d \nabla_\phi \mathcal{L}_{discriminator}^{FGAN}\big(G_\theta,D_\phi,\mathcal{Z},\mathcal{X}\big)
			\end{equation*}
     \EndFor
  \State The generator and discriminator learning rates $\eta_g$ and $\eta_d$ are to be set.
\end{algorithmic}
\end{algorithm}

\section{Experiments}
In our experiments, we tested FGAN on a synthetic 2D dataset to study the training process in FGAN. Next, we tested FGAN for anomaly detection on three datasets: MNIST, CIFAR10 and KDD99, comparing the performance to state-of-the-art anomaly detection methods.

\subsection{2D synthetic dataset}
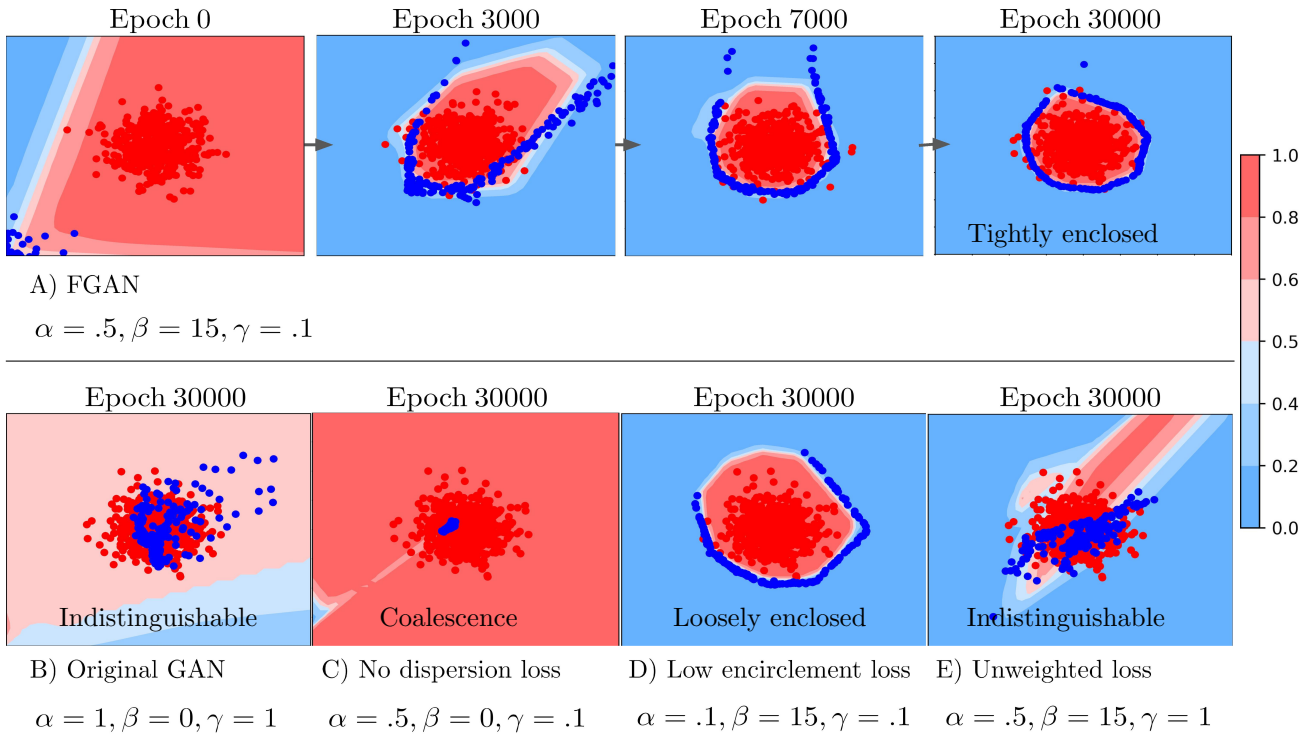
\begin{figure}[H]
\centering
\begin{picture}(800,300)(0,0)
\includegraphics[height = 10cm]{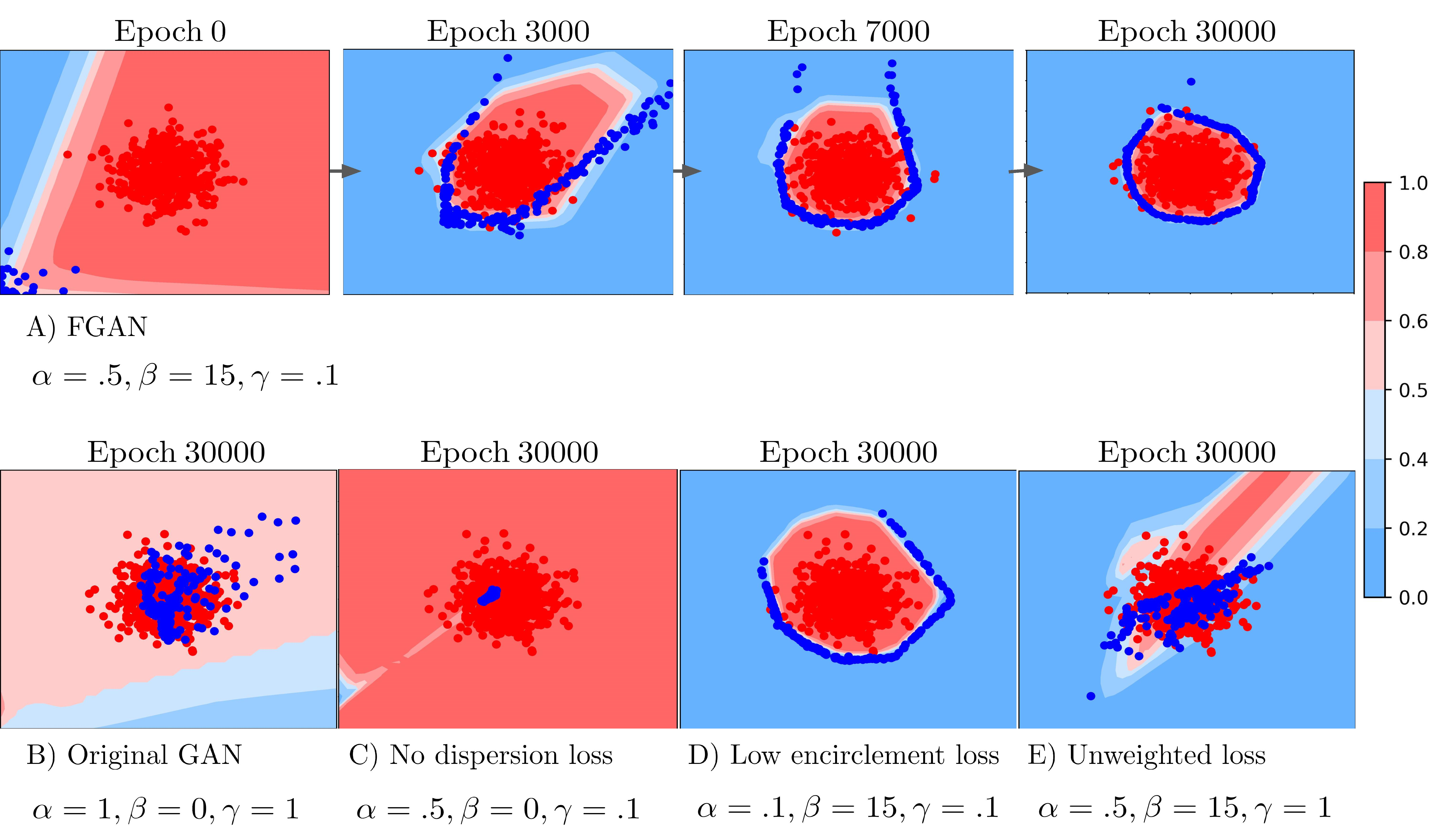}
\put(-125,190){Tightly enclosed}
\put(-465,45){Indistinguishable}
\put(-345,45){Coalescence}
\put(-235,45){Loosely enclosed}
\put(-125,45){Indistinguishable}
\put(-485,145){\line(1,0){460}}
\end{picture}
\caption{FGAN on a normal distribution in two dimensions. Red points are data points and blue points are generated points. The color of the shaded background represents the discriminator score. We trained 5 different models with different hyperparameters over $3\times 10^4$ epochs. (A) shows snapshots of training process for a FGAN trained with optimal hyperparameters, yielding a good discriminator in the end of the training. (B)-(E) are examples of hyperparameters that lead to suboptimal performance of FGAN.}
\label{fig:2Dgauss}
\end{figure}

\noindent
\\We illustrate the effect of FGAN using a 2D synthetic dataset where the data is sampled from a unimodal normal distribution, as shown in Figure \ref{fig:2Dgauss}. The red points represent the real data while the blue points are the generated points. The color of the shaded background represents the discriminator score, where the score increases from blue to red. We trained 5 different FGAN models with different hyperparameters over 30,000 epochs. 

\noindent
\\
In (A), we show snapshots of training process at 4 epochs for an FGAN trained with optimal hyperparameters, yielding a good discriminator at the end of the training. The other examples show hyperparameters that lead to suboptimal performance. (B) shows the result of original GAN \citep{GAN}, where the real data points and generated data points are indistinguishable and the discriminator decision boundary does not surround the real data. (C) illustrates an example of generated points coalescing in one small region and the discriminator classifies most of the data space as positive instances. (D) shows an example of loosely enclosed generated points where the discriminator decision boundary is away from the real data points. (E) shows another example of indistinguishable real data and generated data distributions with bad discriminator decision boundaries. This experiment on the 2D dataset shows that under the optimal hyperparameters, the encirclement loss, dispersion loss and weighted discriminator loss in FGAN give rise to the desired result of the generated samples forming a tight boundary around the dataset.

\subsection{MNIST}
\noindent
\\To show the effectiveness of our proposed idea, we run FGAN for anomaly detection on the MNIST dataset~\citep{mnist}. In each case, we consider data points from a class as `anomalous' (positive class) and data points from the other 9 classes as `normal' (negative class). We then split the entire MNIST dataset which consists of 70000 images from 10 classes into 2 sets as follows: Training set consists of 80\% of all data points in the `normal' class. Testing set consists of the rest 20\% of data in the `normal' class and all data in the `anomalous' class. We then evaluate our model as a binary classifier for normal and anomalous data. Our performance is measured by Area Under Precision and Recall Curve (AUPRC). The architecture as well as hyperparameters to train FGAN are presented in Table \ref{table:archi_MNIST}.

\begin{figure}[H]
\centering
\includegraphics[height = 10cm]{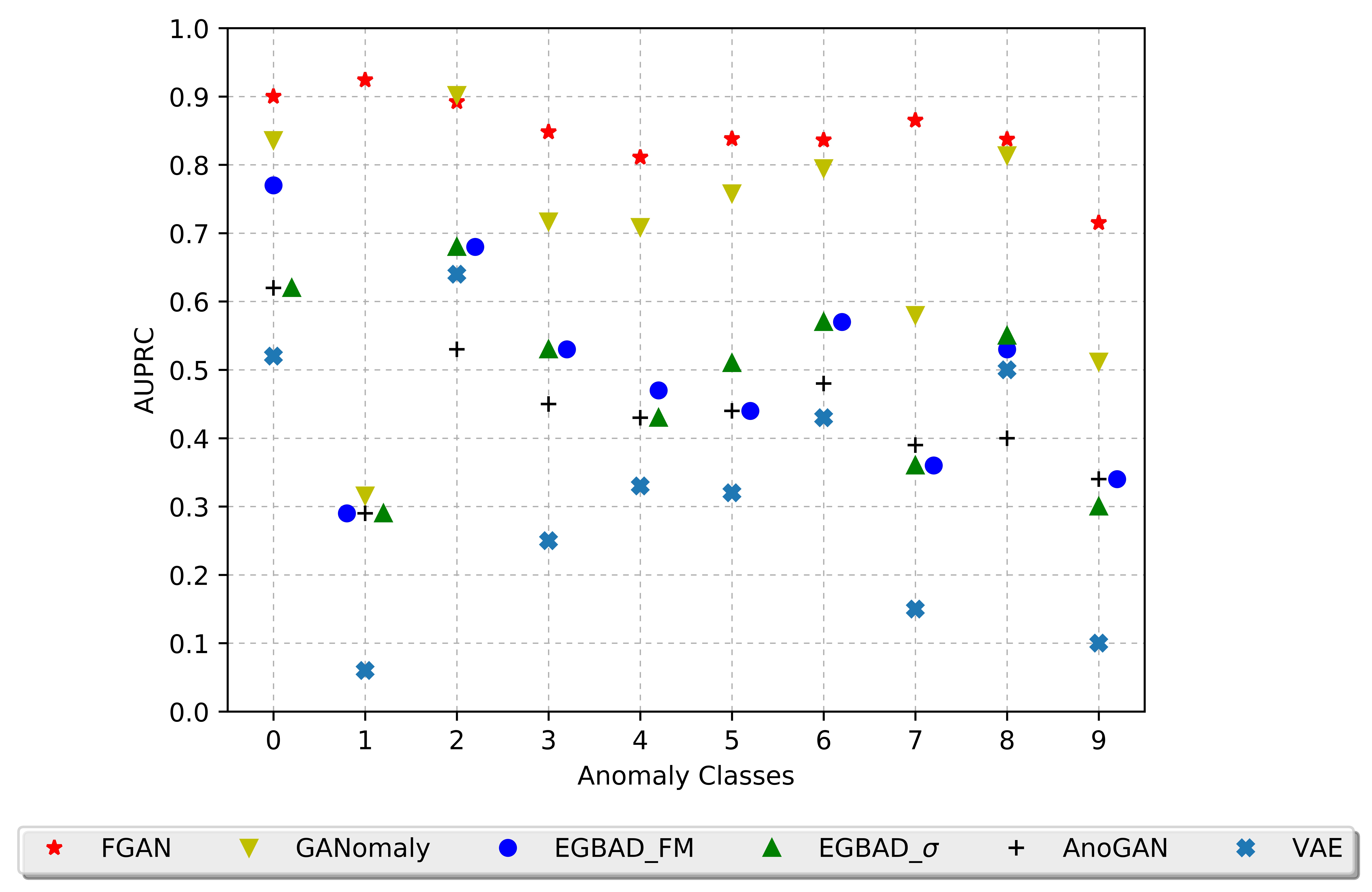}
\caption{Detection accuracies for FGAN and benchmark methods on MNIST dataset.}
\label{fig:result_MNIST}
\end{figure}

\begin{table}[H]
\centering
\begin{tabular}{llllll}
\hline
Operation                   & Kernel   & Strides   & Features Maps/Units   & BN?   & Activation  \\ \hline
\textbf{Generator}                        &          &           &                       &       &                \\
Dense                       &          &           & 1024                  & \checkmark     & ReLU           \\
Dense                       &          &           & 7$\times$7$\times$128               & \checkmark     & ReLU           \\
Transposed Convolution      & 4$\times$4      & 2$\times$2       & 64                    & \checkmark     & ReLU           \\
Transposed Convolution      & 4$\times$4      & 2$\times$2       & 1                     & $\times$     & Tanh           \\ \hline
Latent Dimension            & 200      &           &                       &       &                \\
Encirclement $\alpha$                     & 0.1      &           &                       &       &                \\
Dispersion $\beta$          & 30       &           &                       &       &                \\
Optimizer                   & \multicolumn{5}{l}{Adam(lr=2e-5, decay=1e-4)}                         \\ \hline
\textbf{Discriminator}                        &          &           &                       &       &                \\
Convolution                 & 4$\times$4      & 2$\times$2       & 64                    & $\times$     & Leaky ReLU     \\
Convolution                 & 4$\times$4      & 2$\times$2       & 64                    & $\times$     & Leaky ReLU     \\
Dense                       &          &           & 1024                  & $\times$     & Leaky ReLU     \\
Dense                       &          &           & 1                     & $\times$     & Sigmoid        \\ \hline
Leaky ReLU slope            & 0.1      &           &                       &       &                \\
Optimizer                   & \multicolumn{5}{l}{Adam(lr=1e-5, decay=1e-4)}                         \\
Anomaly $\gamma$              & 0.1      &           &                       &       &                \\ \hline
Epochs                      & 100      &           &                       &       &                \\
Batchsize                   & 200      &           &                       &       &                \\
\end{tabular}
\caption{Architecture and hyperparameters of FGAN on MNIST Dataset}
\label{table:archi_MNIST}
\end{table}

\noindent
\\We train FGAN for 100 epochs with a training batch size of 100 and obtain the average AUPRC across 3 different seeds for each anomalous class. Mean AUPRC in comparison to other benchmark methods are shown in Figure \ref{fig:result_MNIST}. The other benchmark methods are trained using the same setup for splitting of the train and test sets. We reimplement GANomaly \citep{akcay2018ganomaly} using the hyperparameters used by the authors to obtain the AUPRC while the results for EGBAD, AnoGAN and VAE were taken from \citep{EfficientGAN}. As seen from the AUPRC figures, FGAN has the highest accuracy for all but one digit class. Interestingly, for digit classes where the other methods perform badly (eg. digits 1, 7, 9), FGAN's detection accuracy remains high. This shows FGAN's robustness in anomaly detection for the MNIST dataset.

\subsection{CIFAR10}
\noindent
\\Next, we run the anomaly detection test on the CIFAR10 dataset~\citep{krizhevsky2009learning}. Similar to the MNIST experiment, we consider data points from a class as `anomalous' and data points from the other 9 classes as `normal'. We split the entire CIFAR10 dataset with 60000 images into a training set consisting of 80\% of images in the `normal' class and a testing set consisting of the rest 20\% of data in the `normal' class and all data in the `anomalous' class. We train FGAN for 150 epochs with a training batch size of 128. The performance is measured by the Area Under Receiver Operating Characteristics (AUROC) curve, averaged over 3 seeds. The network architecture and hyperparameters to train FGAN are presented in Table \ref{table:archi_CIFAR10}.

\begin{figure}[H]
\centering
\includegraphics[height = 10cm]{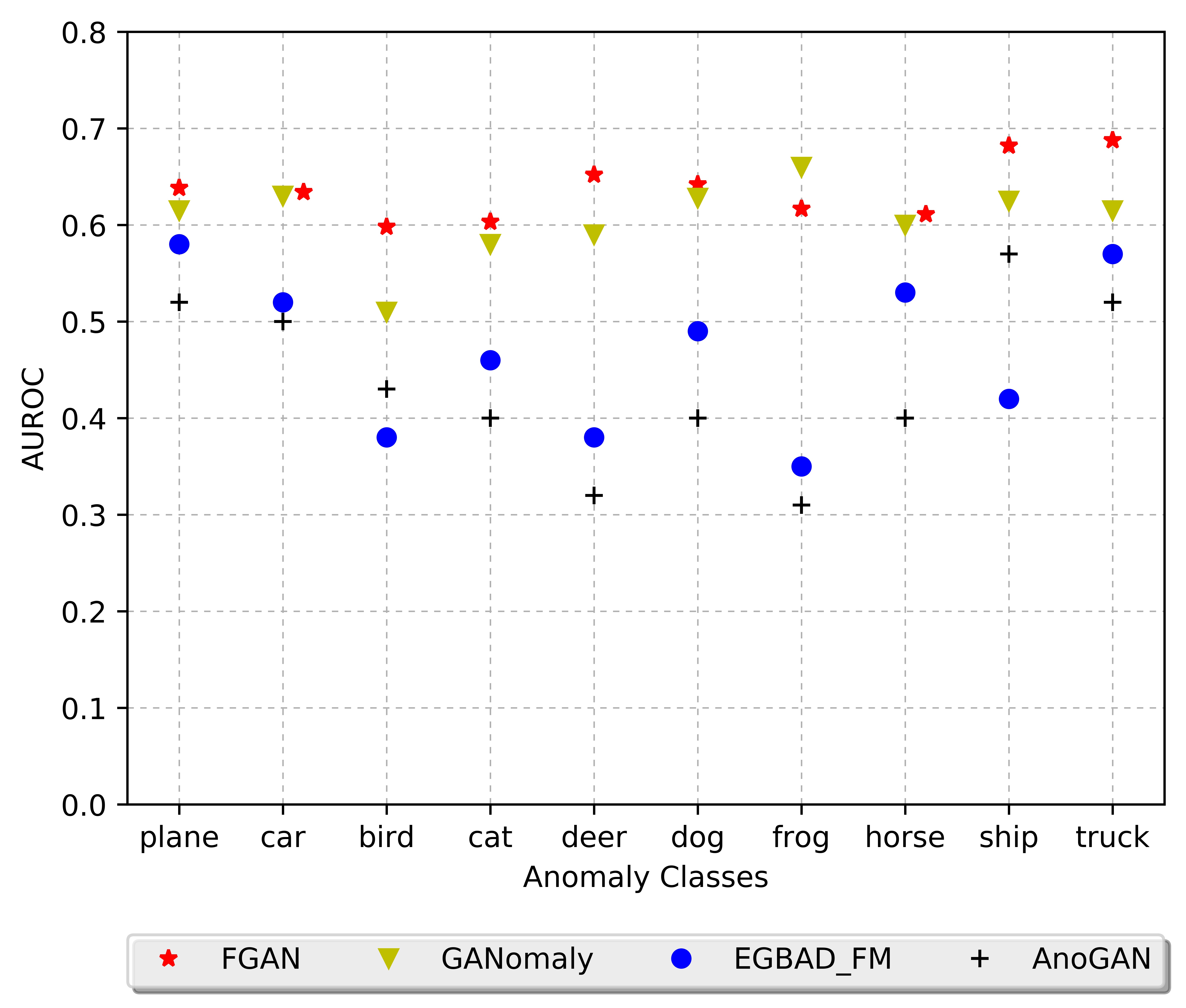}
\caption{Detection accuracies for FGAN and benchmark methods on the CIFAR10 dataset.}
\label{fig:result_CIFAR10}
\end{figure}

\begin{figure}[h!]
    \centering
    \begin{subfigure}[b]{0.3\textwidth}
        \includegraphics[width=\textwidth]{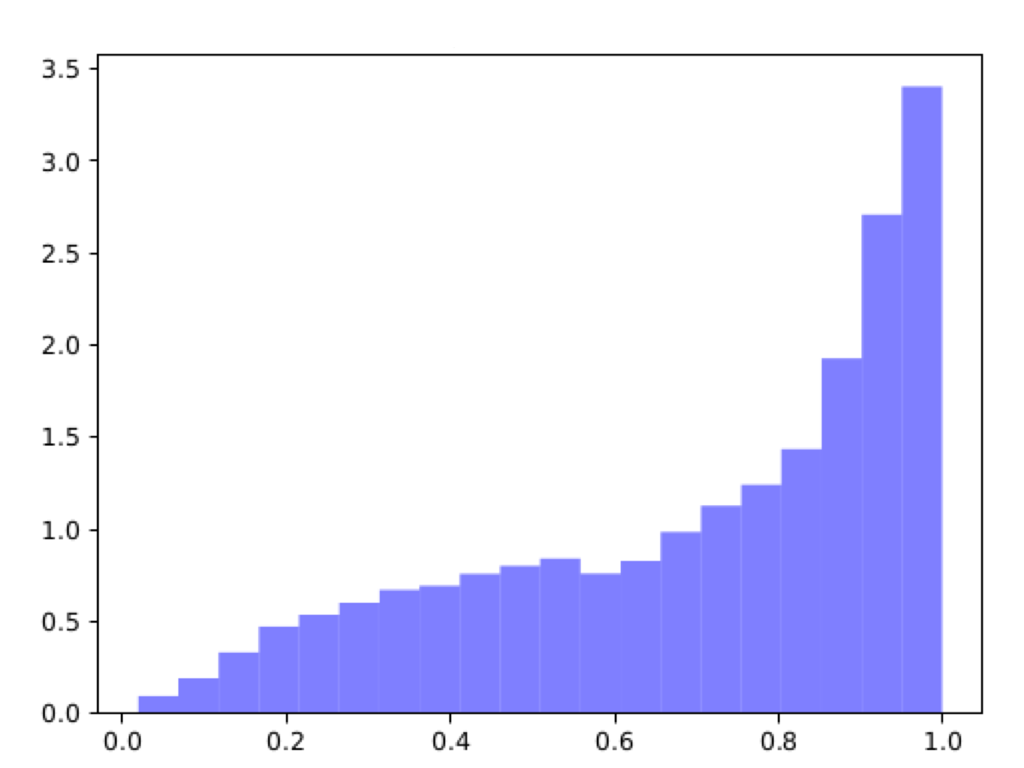}
        \caption{Normal test images}
        \label{fig:score_normal}
    \end{subfigure}
    ~
    \begin{subfigure}[b]{0.3\textwidth}
        \includegraphics[width=\textwidth]{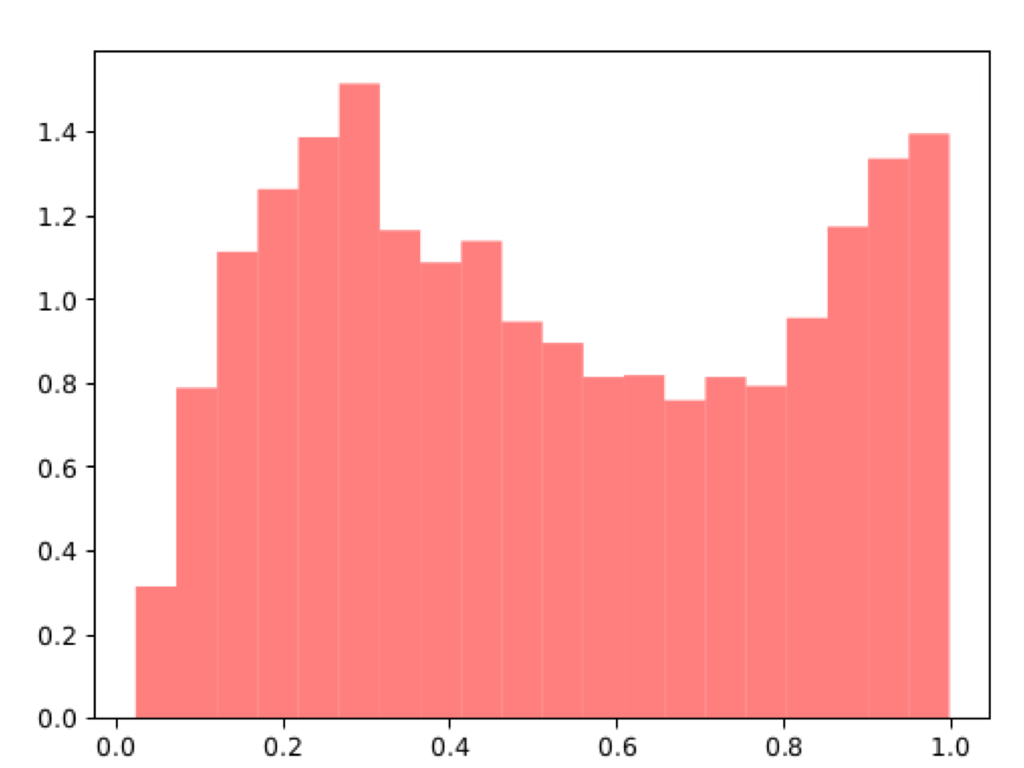}
        \caption{Anomalous images}
        \label{fig:score_ano}
    \end{subfigure}
    ~
    \begin{subfigure}[b]{0.3\textwidth}
        \includegraphics[width=\textwidth]{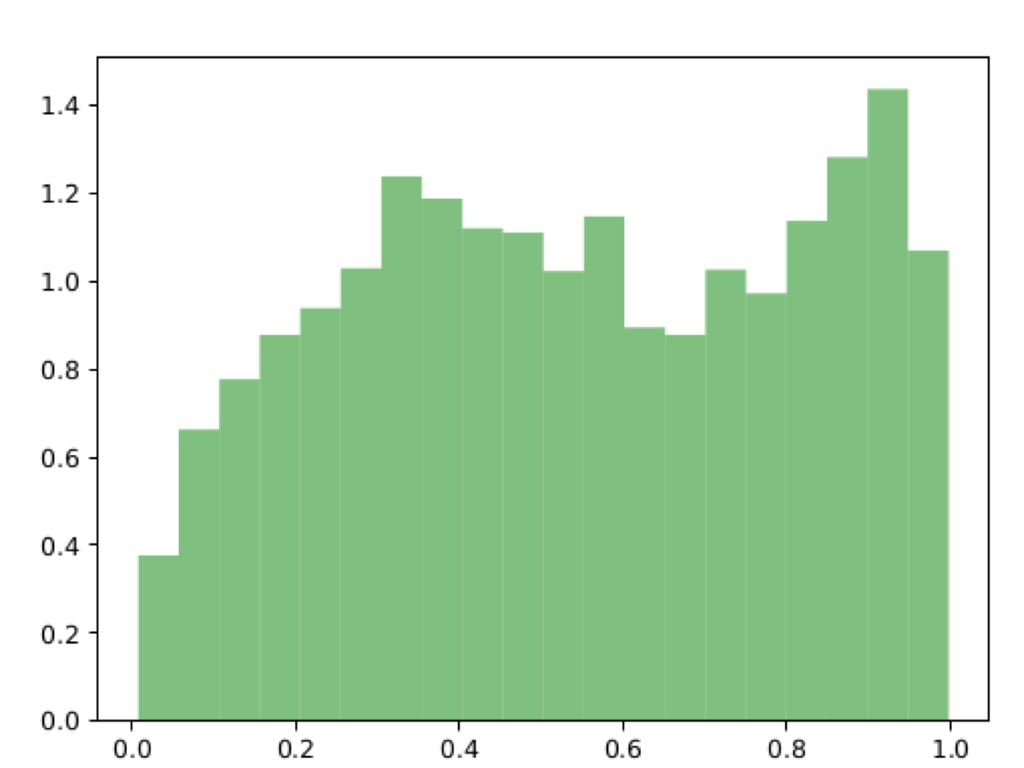}
        \caption{Generated images}
        \label{fig:score_gen}
    \end{subfigure}
    \caption{CIFAR10 anomalous class: ship. Histogram of discriminator scores for (a) normal test images, (b) anomalous images and (c) generated images. The normal test images have a distribution that is skewed towards higher scores, which is expected from the FGAN loss function. The distributions for the anomalous and generated images are bimodal.}\label{fig:CIFAR10_ship_scores}
\end{figure}

\begin{figure}[H]
\centering
\includegraphics[width=0.8\columnwidth]{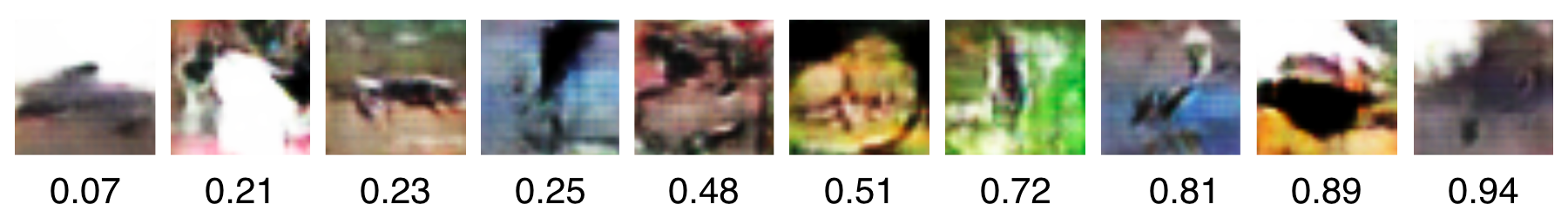}
\caption{CIFAR10 anomalous class: ship. Examples of generated images shown in increasing order of the discriminator score.}
\label{fig:ship_generated_images}
\end{figure}

\begin{table}[H]
\centering
\begin{tabular}{@{}llllll@{}}
\hline
Operation & Kernel & Strides & Feature Maps / Units & BN? & Activation \\ \midrule
\textbf{Generator} &  &  &  &  &  \\
Dense &  &  & 2$\times$2$\times$256 & $\checkmark$ & Leaky ReLU \\
Transposed Convolution & 5$\times$5 & 2$\times$2 & 128 & $\checkmark$ & Leaky ReLU \\
Transposed Convolution & 5$\times$5 & 2$\times$2 & 64 & $\checkmark$ & Leaky ReLU \\
Transposed Convolution & 5$\times$5 & 2$\times$2 & 32 & $\checkmark$ & Leaky ReLU \\
Transposed Convolution & 5$\times$5 & 2$\times$2 & 3 & $\times$ & Tanh \\ \midrule
Latent Dimension & \multicolumn{5}{l}{256} \\
Leaky ReLU Slope & \multicolumn{5}{l}{0.2} \\
Encirclement $\alpha$ & \multicolumn{5}{l}{0.5} \\
Dispersion $\beta$ & \multicolumn{5}{l}{10} \\
Optimizer & \multicolumn{5}{l}{Adam(lr=1e-3, beta\_1 = 0.5, beta\_2 = 0.999, decay=1e-5)} \\ \midrule
\textbf{Discriminator} &  &  &  &  &  \\
Convolution & 5$\times$5 & 2$\times$2 & 32 & $\checkmark$ & Leaky ReLU \\
Convolution & 5$\times$5 & 2$\times$2 & 64 & $\checkmark$ & Leaky ReLU \\
Convolution & 5$\times$5 & 2$\times$2 & 128 & $\checkmark$ & Leaky ReLU \\
Convolution & 5$\times$5 & 2$\times$2 & 256 & $\times$ & Leaky ReLU \\
Dropout &  &  & 0.2 & $\times$ &  \\
Dense &  &  & 1 & $\times$ & Sigmoid \\ \midrule
Leaky ReLU Slope & 0.2 &  &  &  &  \\
Weight Decay & 0.5 &  &  &  &  \\
Optimizer & \multicolumn{5}{l}{Adam(lr=1e-4, beta\_1 = 0.5, beta\_2 = 0.999, decay=1e-5)} \\
Anomaly $\gamma$ & 0.5 &  &  &  &  \\ \midrule
Epochs & 150 &  &  &  &  \\
Batch Size & 128 &  &  &  & 
\end{tabular}
\caption{Architecture and hyperparameters of FGAN on CIFAR-10 Dataset}
\label{table:archi_CIFAR10}
\end{table}

\noindent
\\The anomaly detection results are shown in Figure \ref{fig:result_CIFAR10}. For all but one anomaly class, FGAN's accuracy is the highest among all methods, with AUROC of at least 60\% across all classes. For the challenging `bird' class where the state-of-the-art GANomaly method has an AUROC of just above 50\%, FGAN manages an AUROC of 60\%, showing its detection robustness.

\noindent
\\
In Figure \ref{fig:CIFAR10_ship_scores}, we analyze the distribution of discriminator scores for the normal test images, anomalous images and generated images where the anomalous class is `ship'. The normal test images scores are skewed towards the high score of 1.0 which is expected from the FGAN loss function. The anomalous images scores show a bimodal distribution with modes at 0.3 and 1.0, which means some anomalous images are challenging to detect. This explains the relatively good anomaly detection results in Figure \ref{fig:result_CIFAR10}. The distribution of scores for generated images is spread across the entire range which means the generated samples do not converge at a score of $\alpha=0.5$, although this has not significantly impacted anomaly detection performance.

\noindent
\\
Table \ref{table:CIFAR10_avgscores} shows the average discriminator score for the anomalous `ship' class, the other 9 classes and the generated images. All scores are taken from the test set. Interestingly the `airplane' class has a lower score than the anomalous `ship' class, and this may indicate that these two classes are semantically similar and challenging to differentiate compared to other classes. In Figure \ref{fig:ship_generated_images}, we show examples of the generated images with their corresponding discriminator scores. The generated images are relatively realistic, resembling natural images, though there is no identifiable pattern from the discriminator scores.

\begin{table}[h!]
\begin{tabular}{@{}cccccccccccc@{}}
\toprule
Class                                                                  & Airplane & Ship  & \begin{tabular}[c]{@{}c@{}}Generated \\ images\end{tabular} & Dog   & Bird  & Cat   & Car   & Truck & Deer & Horse & Frog  \\ \midrule
\begin{tabular}[c]{@{}c@{}}Average \\ score\end{tabular} & 0.505    & 0.521 & 0.543                                                       & 0.669 & 0.671 & 0.686 & 0.717 & 0.737 & 0.75 & 0.752 & 0.795 \\ \bottomrule
\end{tabular}
\caption{CIFAR10 anomalous class: ship. Average discriminator scores for the anomalous `ship' class, the other 9 classes and the generated images. All scores were taken from the test set and the average scores are shown in increasing order.}
\label{table:CIFAR10_avgscores}
\end{table}

\subsection{KDD99}
\noindent
\\In order to further validate the merits of our approach, we test FGAN on the KDDCUP99 10 percent dataset~\citep{kdd99}. We follow the experimental setup of \citep{DAGMM} \citep{EfficientGAN} in our experiments. In the KDD99 dataset, data is stratified into the `non-attack' class and other classes with various attacks. We lump all the other classes with various attacks as one class and call it the `attack' class. We then train FGAN on the `attack' class only because the proportion of data belonging to the `attack' class is much larger than the proportion of data belonging to the `non-attack' class. The objective is to detect the `non-attack' instances.

\noindent
\\We split the KDD99 dataset as follows:
Training set consists of 50\% of all data points in the `attack' class.
Testing set consists of the remaining 50\% of data in the `attack' class and 50\% of data in the `non-attack' class. We then evaluate our approach against Efficient-GAN. The architecture and hyperparameters to train FGAN for KDD99 dataset is shown in Table \ref{table:archi_KDD}. 

\begin{table}[H]
\centering
\begin{tabular}{@{}lllll@{}}
\toprule
Operation        & Units & Non Linearity & Dropout & L2 Regularization \\ \midrule
\textbf{Generator}             &       &               & 0       & 0                 \\
Dense            & 64    & ReLU          & 0.2     & 0                  \\
Dense            & 128   & ReLU          & 0.2     & 0                 \\
Dense (output)   & 121   & Linear        & 0       & 0                \\\midrule
Latent Dimension & \multicolumn{4}{l}{32}                              \\
Encirclement $\alpha$		     & \multicolumn{4}{l}{0.5}                             \\
Dispersion $\beta$ & \multicolumn{4}{l}{30}                              \\ 
Optimizer        & \multicolumn{4}{l}{Adam(lr = 1e-4, decay = 1e-3)}   \\\midrule
\textbf{Discriminator}             &       &               & 0        &                   \\
Dense            & 256   & Leaky ReLU    & 0       & 0                 \\
Dense            & 128   & Leaky ReLU    & 0       & 0                 \\
Dense            & 128   & Leaky ReLU    & 0       & 0               \\
Dense (output)   & 1     & Sigmoid       & 0        &                   \\\midrule
Leaky ReLU Slope & \multicolumn{4}{l}{0.1} 							   \\
Optimizer        & \multicolumn{4}{l}{SGD(lr = 8e-6, decay = 1e-3)}   \\
Anomaly $\gamma$   & \multicolumn{4}{l}{0.5}   \\\bottomrule                      
\end{tabular}
\caption{Architecture and hyperparameters for FGAN on KDD99 Dataset.}
\label{table:archi_KDD}
\end{table}

\noindent
\\ We train with a batch size of 256 for both discriminator and generator for 50 epochs. The precision, recall and F1 scores are averaged over 10 different consecutive seeds, as shown in Table \ref{table:results_KDD}. FGAN has the best anomaly detection accuracies compared to the other benchmark methods.

\begin{table}[H]
\centering
\begin{tabular}{|l|l|l|l|}
\textbf{Model}                & \textbf{Precision}      & \textbf{Recall}         & \textbf{F1}             \\ \hline
OC-SVM               & 0.7457         & 0.8523         & 0.7954         \\
DSEBM-r              & 0.8521         & 0.6472         & 0.7328         \\
DSEBM-e              & 0.8619         & 0.6446         & 0.7399         \\
DAGMM-NVI            & 0.9290         & 0.9447         & 0.9368         \\
DAGMM                & 0.9297         & 0.9442         & 0.9369         \\
AnoGANfm             & $0.88\pm 3\times 10^{-2}$ & $0.83\pm 3\times 10^{-2}$ & $0.89\pm 3\times 10^{-2}$ \\
AnoGANsigmoid        & $0.8\pm 0.1$ & $0.8\pm 0.1$ & $0.8\pm 0.1$ \\
Efficient-GANfm      & $0.9\pm 0.1$ & $0.95\pm 2\times 10^{-2}$ & $0.91\pm 7\times 10^{-2}$ \\
Efficient-GANsigmoid & $0.92\pm 7\times 10^{-2}$ & $0.96\pm 1\times 10^{-2}$  & $0.94\pm 4\times 10^{-2}$ \\
FGAN                 & \textbf{0.954}$\pm 9\times 10^{-3}$ & \textbf{0.969}$\pm 9\times 10^{-3}$ & \textbf{0.95}$\pm 2 \times 10^{-2}$
\end{tabular}
\caption{Performance on the KDD99 dataset. Values for OC-SVM, DSEBM, DAGMM were obtained from \citep{DSEBM}, \citep{DAGMM}. Values for AnoGAN and Efficient-GAN were obtained from \citep{EfficientGAN}. Precision, Recall, and F1 Score are calculated with `non-attack' class being the positive class and `attack' class being the negative class.}
\label{table:results_KDD}
\end{table}

\section{Discussion}
%1. anomaly detection is important task
%2. current state-of-the-art methods on anomaly detection for complex high-dim images use GAN
%3. But normal GAN objective encourages distribution of generated samples to overlap with real data, which is not %directly aligned with the anomaly detection objective, and hence people find the resulting discriminator is ineffective for anomaly detector
%4. Our contribution: We propose a simple tweak to the GAN objective such that generated samples lie at the boundary of real data distribution, then use the discriminator score as anomaly score
%5. Experiments show improvement over current best GAN-based methods
\noindent
\\
State-of-the-art anomaly detection methods for complex high-dimensional data are based on generative adversarial networks. However, in this paper, we identify that the usual GAN loss objective is not directly aligned with the anomaly detection objective: the loss encourages the distribution of generated samples to overlap with real data. Hence, the resulting discriminator has been found to be ineffective for anomaly detector. Hence, we propose simple modifications to the GAN loss such that the generated samples lie at the boundary of real data distribution. Our method, called Fence GAN, uses the discriminator score as anomaly score. With the modified GAN loss, Fence GAN does not need to rely on reconstruction loss from the generator and does not require modifications to the basic GAN architecture unlike Efficient GAN and GANomaly. 

\noindent
\\
On the MNIST, CIFAR10 and KDD99 datasets, Fence GAN outperforms existing methods in anomaly detection. We have shown that with simple modifications to the GAN loss, the basic GAN architecture and training scheme can produce an effective anomaly detector for complex high-dimensional data.

\bibliographystyle{plainnat}
\bibliography{Paper}

\begin{thebibliography}{35}
\providecommand{\natexlab}[1]{#1}
\providecommand{\url}[1]{\texttt{#1}}
\expandafter\ifx\csname urlstyle\endcsname\relax
  \providecommand{\doi}[1]{doi: #1}\else
  \providecommand{\doi}{doi: \begingroup \urlstyle{rm}\Url}\fi

\bibitem[Akcay et~al.(2018)Akcay, Atapour-Abarghouei, and
  Breckon]{akcay2018ganomaly}
Samet Akcay, Amir Atapour-Abarghouei, and Toby~P Breckon.
\newblock Ganomaly: Semi-supervised anomaly detection via adversarial training.
\newblock \emph{arXiv preprint arXiv:1805.06725}, 2018.

\bibitem[An and Cho(2015)]{an2015variational}
Jinwon An and Sungzoon Cho.
\newblock Variational autoencoder based anomaly detection using reconstruction
  probability.
\newblock \emph{Special Lecture on IE}, 2:\penalty0 1--18, 2015.

\bibitem[Bay et~al.(2000)Bay, Kibler, Pazzani, and Smyth]{kdd99}
Stephen~D Bay, Dennis~F Kibler, Michael~J Pazzani, and Padhraic Smyth.
\newblock The uci kdd archive of large data sets for data mining research and
  experimentation.
\newblock \emph{SIGKDD Explorations}, 2:\penalty0 81, 2000.

\bibitem[Beggel et~al.(2019)Beggel, Pfeiffer, and Bischl]{beggel2019robust}
Laura Beggel, Michael Pfeiffer, and Bernd Bischl.
\newblock Robust anomaly detection in images using adversarial autoencoders.
\newblock \emph{arXiv preprint arXiv:1901.06355}, 2019.

\bibitem[Chandola et~al.(2009)Chandola, Banerjee, and
  Kumar]{chandola2009anomaly}
Varun Chandola, Arindam Banerjee, and Vipin Kumar.
\newblock Anomaly detection: A survey.
\newblock \emph{ACM computing surveys (CSUR)}, 41\penalty0 (3):\penalty0 15,
  2009.

\bibitem[Dai et~al.(2017)Dai, Yang, Yang, Cohen, and
  Salakhutdinov]{dai2017good}
Zihang Dai, Zhilin Yang, Fan Yang, William~W Cohen, and Ruslan~R Salakhutdinov.
\newblock Good semi-supervised learning that requires a bad gan.
\newblock In \emph{Advances in Neural Information Processing Systems}, pages
  6510--6520, 2017.

\bibitem[Deecke et~al.(2018)Deecke, Vandermeulen, Ruff, Mandt, and
  Kloft]{deecke2018anomaly}
Lucas Deecke, Robert Vandermeulen, Lukas Ruff, Stephan Mandt, and Marius Kloft.
\newblock Anomaly detection with generative adversarial networks.
\newblock 2018.

\bibitem[Erfani et~al.(2016)Erfani, Rajasegarar, Karunasekera, and
  Leckie]{erfani2016high}
Sarah~M Erfani, Sutharshan Rajasegarar, Shanika Karunasekera, and Christopher
  Leckie.
\newblock High-dimensional and large-scale anomaly detection using a linear
  one-class svm with deep learning.
\newblock \emph{Pattern Recognition}, 58:\penalty0 121--134, 2016.

\bibitem[Eskin et~al.(2002)Eskin, Arnold, Prerau, Portnoy, and
  Stolfo]{eskin2002geometric}
Eleazar Eskin, Andrew Arnold, Michael Prerau, Leonid Portnoy, and Sal Stolfo.
\newblock A geometric framework for unsupervised anomaly detection.
\newblock In \emph{Applications of data mining in computer security}, pages
  77--101. Springer, 2002.

\bibitem[Garcia-Teodoro et~al.(2009)Garcia-Teodoro, Diaz-Verdejo,
  Maci{\'a}-Fern{\'a}ndez, and V{\'a}zquez]{garcia2009anomaly}
Pedro Garcia-Teodoro, Jesus Diaz-Verdejo, Gabriel Maci{\'a}-Fern{\'a}ndez, and
  Enrique V{\'a}zquez.
\newblock Anomaly-based network intrusion detection: Techniques, systems and
  challenges.
\newblock \emph{computers \& security}, 28\penalty0 (1-2):\penalty0 18--28,
  2009.

\bibitem[Goodfellow et~al.(2014{\natexlab{a}})Goodfellow, Pouget-Abadie, Mirza,
  Xu, Warde-Farley, Ozair, Courville, and Bengio]{goodfellow2014generative}
Ian Goodfellow, Jean Pouget-Abadie, Mehdi Mirza, Bing Xu, David Warde-Farley,
  Sherjil Ozair, Aaron Courville, and Yoshua Bengio.
\newblock Generative adversarial nets.
\newblock In \emph{Advances in neural information processing systems}, pages
  2672--2680, 2014{\natexlab{a}}.

\bibitem[Goodfellow et~al.(2014{\natexlab{b}})Goodfellow, Pouget-Abadie, Mirza,
  Xu, Warde-Farley, Ozair, Courville, and Bengio]{GAN}
Ian~J. Goodfellow, Jean Pouget-Abadie, Mehdi Mirza, Bing Xu, David
  Warde-Farley, Sherjil Ozair, Aaron~C. Courville, and Yoshua Bengio.
\newblock Generative adversarial nets.
\newblock In \emph{NIPS}, 2014{\natexlab{b}}.

\bibitem[G{\"o}rnitz et~al.(2015)G{\"o}rnitz, Braun, and
  Kloft]{gornitz2015hidden}
Nico G{\"o}rnitz, Mikio Braun, and Marius Kloft.
\newblock Hidden markov anomaly detection.
\newblock In \emph{International Conference on Machine Learning}, pages
  1833--1842, 2015.

\bibitem[Hawkins(1980)]{hawkins1980identification}
Douglas~M Hawkins.
\newblock \emph{Identification of outliers}, volume~11.
\newblock Springer, 1980.

\bibitem[He et~al.(2016)He, Zhang, Ren, and Sun]{he2016deep}
Kaiming He, Xiangyu Zhang, Shaoqing Ren, and Jian Sun.
\newblock Deep residual learning for image recognition.
\newblock In \emph{Proceedings of the IEEE conference on computer vision and
  pattern recognition}, pages 770--778, 2016.

\bibitem[Krizhevsky and Hinton(2009)]{krizhevsky2009learning}
Alex Krizhevsky and Geoffrey Hinton.
\newblock Learning multiple layers of features from tiny images.
\newblock Technical report, Citeseer, 2009.

\bibitem[LeCun and Cortes(2010)]{mnist}
Yann LeCun and Corinna Cortes.
\newblock {MNIST} handwritten digit database.
\newblock 2010.
\newblock URL \url{http://yann.lecun.com/exdb/mnist/}.

\bibitem[Leveau and Joly(2017)]{leveau2017adversarial}
Valentin Leveau and Alexis Joly.
\newblock Adversarial autoencoders for novelty detection.
\newblock 2017.

\bibitem[Lim et~al.(2018)Lim, Loo, Tran, Cheung, Roig, and
  Elovici]{lim2018doping}
Swee~Kiat Lim, Yi~Loo, Ngoc-Trung Tran, Ngai-Man Cheung, Gemma Roig, and Yuval
  Elovici.
\newblock Doping: Generative data augmentation for unsupervised anomaly
  detection with gan.
\newblock In \emph{2018 IEEE International Conference on Data Mining (ICDM)},
  pages 1122--1127. IEEE, 2018.

\bibitem[Mahadevan et~al.(2010)Mahadevan, Li, Bhalodia, and
  Vasconcelos]{mahadevan2010anomaly}
Vijay Mahadevan, Weixin Li, Viral Bhalodia, and Nuno Vasconcelos.
\newblock Anomaly detection in crowded scenes.
\newblock In \emph{Computer Vision and Pattern Recognition (CVPR), 2010 IEEE
  Conference on}, pages 1975--1981. IEEE, 2010.

\bibitem[Nicolau et~al.(2016)Nicolau, McDermott, et~al.]{nicolau2016one}
Miguel Nicolau, James McDermott, et~al.
\newblock One-class classification for anomaly detection with kernel density
  estimation and genetic programming.
\newblock In \emph{European Conference on Genetic Programming}, pages 3--18.
  Springer, 2016.

\bibitem[Radford et~al.(2015)Radford, Metz, and
  Chintala]{radford2015unsupervised}
Alec Radford, Luke Metz, and Soumith Chintala.
\newblock Unsupervised representation learning with deep convolutional
  generative adversarial networks.
\newblock \emph{arXiv preprint arXiv:1511.06434}, 2015.

\bibitem[Ravanbakhsh et~al.(2017)Ravanbakhsh, Sangineto, Nabi, and
  Sebe]{ravanbakhsh2017training}
Mahdyar Ravanbakhsh, Enver Sangineto, Moin Nabi, and Nicu Sebe.
\newblock Training adversarial discriminators for cross-channel abnormal event
  detection in crowds.
\newblock \emph{arXiv preprint arXiv:1706.07680}, 2017.

\bibitem[Ronneberger et~al.(2015)Ronneberger, Fischer, and
  Brox]{ronneberger2015u}
Olaf Ronneberger, Philipp Fischer, and Thomas Brox.
\newblock U-net: Convolutional networks for biomedical image segmentation.
\newblock In \emph{International Conference on Medical image computing and
  computer-assisted intervention}, pages 234--241. Springer, 2015.

\bibitem[Schlegl et~al.(2017)Schlegl, Seeb{\"o}ck, Waldstein, Schmidt-Erfurth,
  and Langs]{schlegl2017unsupervised}
Thomas Schlegl, Philipp Seeb{\"o}ck, Sebastian~M Waldstein, Ursula
  Schmidt-Erfurth, and Georg Langs.
\newblock Unsupervised anomaly detection with generative adversarial networks
  to guide marker discovery.
\newblock In \emph{International Conference on Information Processing in
  Medical Imaging}, pages 146--157. Springer, 2017.

\bibitem[Sch{\"o}lkopf et~al.(2001)Sch{\"o}lkopf, Platt, Shawe-Taylor, Smola,
  and Williamson]{scholkopf2001estimating}
Bernhard Sch{\"o}lkopf, John~C Platt, John Shawe-Taylor, Alex~J Smola, and
  Robert~C Williamson.
\newblock Estimating the support of a high-dimensional distribution.
\newblock \emph{Neural computation}, 13\penalty0 (7):\penalty0 1443--1471,
  2001.

\bibitem[Seidel(1986)]{convexhull}
Raimund Seidel.
\newblock Constructing higher-dimensional convex hulls at logarithmic cost per
  face.
\newblock pages 404--413, 01 1986.
\newblock \doi{10.1145/12130.12172}.

\bibitem[Smith et~al.(2002)Smith, Bivens, Embrechts, Palagiri, and
  Szymanski]{smith2002clustering}
Rasheda Smith, Alan Bivens, Mark Embrechts, Chandrika Palagiri, and Boleslaw
  Szymanski.
\newblock Clustering approaches for anomaly based intrusion detection.
\newblock \emph{Proceedings of intelligent engineering systems through
  artificial neural networks}, pages 579--584, 2002.

\bibitem[Srivastava et~al.(2008)Srivastava, Kundu, Sural, and
  Majumdar]{srivastava2008credit}
Abhinav Srivastava, Amlan Kundu, Shamik Sural, and Arun Majumdar.
\newblock Credit card fraud detection using hidden markov model.
\newblock \emph{IEEE Transactions on dependable and secure computing},
  5\penalty0 (1):\penalty0 37--48, 2008.

\bibitem[Xu et~al.(2018)Xu, Chen, Zhao, Li, Bu, Li, Liu, Zhao, Pei, Feng,
  et~al.]{xu2018unsupervised}
Haowen Xu, Wenxiao Chen, Nengwen Zhao, Zeyan Li, Jiahao Bu, Zhihan Li, Ying
  Liu, Youjian Zhao, Dan Pei, Yang Feng, et~al.
\newblock Unsupervised anomaly detection via variational auto-encoder for
  seasonal kpis in web applications.
\newblock In \emph{Proceedings of the 2018 World Wide Web Conference on World
  Wide Web}, pages 187--196. International World Wide Web Conferences Steering
  Committee, 2018.

\bibitem[{Zenati} et~al.(2018){Zenati}, {Foo}, {Lecouat}, {Manek}, and
  {Ramaseshan Chandrasekhar}]{EfficientGAN}
H.~{Zenati}, C.~S. {Foo}, B.~{Lecouat}, G.~{Manek}, and V.~{Ramaseshan
  Chandrasekhar}.
\newblock {Efficient GAN-Based Anomaly Detection}.
\newblock \emph{ArXiv e-prints}, February 2018.

\bibitem[Zenati et~al.(2018)Zenati, Foo, Lecouat, Manek, and
  Chandrasekhar]{zenati2018efficient}
Houssam Zenati, Chuan~Sheng Foo, Bruno Lecouat, Gaurav Manek, and
  Vijay~Ramaseshan Chandrasekhar.
\newblock Efficient gan-based anomaly detection.
\newblock \emph{arXiv preprint arXiv:1802.06222}, 2018.

\bibitem[Zhai et~al.(2016{\natexlab{a}})Zhai, Cheng, Lu, and Zhang]{DSEBM}
Shuangfei Zhai, Yu~Cheng, Weining Lu, and Zhongfei Zhang.
\newblock Deep structured energy based models for anomaly detection.
\newblock \emph{CoRR}, abs/1605.07717, 2016{\natexlab{a}}.
\newblock URL \url{http://arxiv.org/abs/1605.07717}.

\bibitem[Zhai et~al.(2016{\natexlab{b}})Zhai, Cheng, Lu, and
  Zhang]{zhai2016deep}
Shuangfei Zhai, Yu~Cheng, Weining Lu, and Zhongfei Zhang.
\newblock Deep structured energy based models for anomaly detection.
\newblock \emph{arXiv preprint arXiv:1605.07717}, 2016{\natexlab{b}}.

\bibitem[Zong et~al.(2018)Zong, Song, Min, Cheng, Lumezanu, Cho, and
  Chen]{DAGMM}
Bo~Zong, Qi~Song, Martin~Renqiang Min, Wei Cheng, Cristian Lumezanu, Daeki Cho,
  and Haifeng Chen.
\newblock Deep autoencoding gaussian mixture model for unsupervised anomaly
  detection.
\newblock In \emph{International Conference on Learning Representations}, 2018.
\newblock URL \url{https://openreview.net/forum?id=BJJLHbb0-}.

\end{thebibliography}
\end{document}